\documentclass[journal]{IEEEtran}
\usepackage{graphicx}
\usepackage{multirow} 
\ifCLASSINFOpdf
\else
\fi
\begin{document}
%
\title{Real-Time On-Drone Hyperspectral Anomaly Detection for Wilderness Search and Rescue}
%
%
%

\author{Jacqueline Liu,
        Gyabaah Kyere,
        Chengyan Liu,
        Kenneth Chiu,
        Lynn Marie (Kic'i) Church,
        Sean Gleason,
        and Jayson Boubin%
\thanks{Jacqueline Liu, Chengyan Liu, Kenneth Chiu, and Jayson Boubin are with the School of Computing, Binghamton University, Binghamton, NY 13902 USA (email: jliu28@binghamton.edu; cliu118@binghamton.edu; kchiu@binghamton.edu; jboubin@binghamton.edu)}%
\thanks{Gyabaah Kyere and Sean Gleason are with the Mathematics and Computer Science Department, Hampden-Sydney College, Hampden-Sydney, VA 23943 USA (email: gyabaakyere@gmail.com; sgleason@nalaquq.com)}%
\thanks{Lynn Marie (Kic'i) Church and Sean Gleason are with Nalaquq, LLC, Quinhagak, AK 99655 USA (email:lchurch@nalaquq.com; sgleason@nalaquq.com)}}

\maketitle

\begin{abstract}
Wilderness search-and-rescue (WSAR) teams use uncrewed aerial systems (UASs) to survey terrain for indicators of missing persons that are rare, blend into the background, or lack thermal signatures. Hyperspectral imaging (HSI) provides rich spectral information, but its high data volume challenges real-time on-drone processing. We present SABL, an edge-computing pipeline for on-drone \underline{S}pectral \underline{A}nomaly-\underline{B}ased \underline{L}ocalization. SABL integrates hyperspectral streaming with exponentially moving Reed–Xiaoli (ERX) anomaly detection and DBSCAN spatial filtering to group pixel-level anomalies into candidate search regions. SABL is evaluated on data collected during a 10-day deployment with Quinhagak Search and Rescue (QSAR) in remote Alaska. It processes hyperspectral scanlines at 5.67-5.73 ms per line, meeting the camera’s 10-ms acquisition interval; DBSCAN requires only 6.5-12.5 ms per 100-line update for filtering. SABL achieves approximately 56\% and 99\% object-level precision on QSAR-selected search indicators across two scenes, with corresponding recall of 55\% and 29\%, providing significant background filtering while identifying difficult-to-see anomalous search indicators.
\end{abstract}


\IEEEpeerreviewmaketitle

\begin{figure*}[ht]
\centering
\includegraphics[scale=0.6]{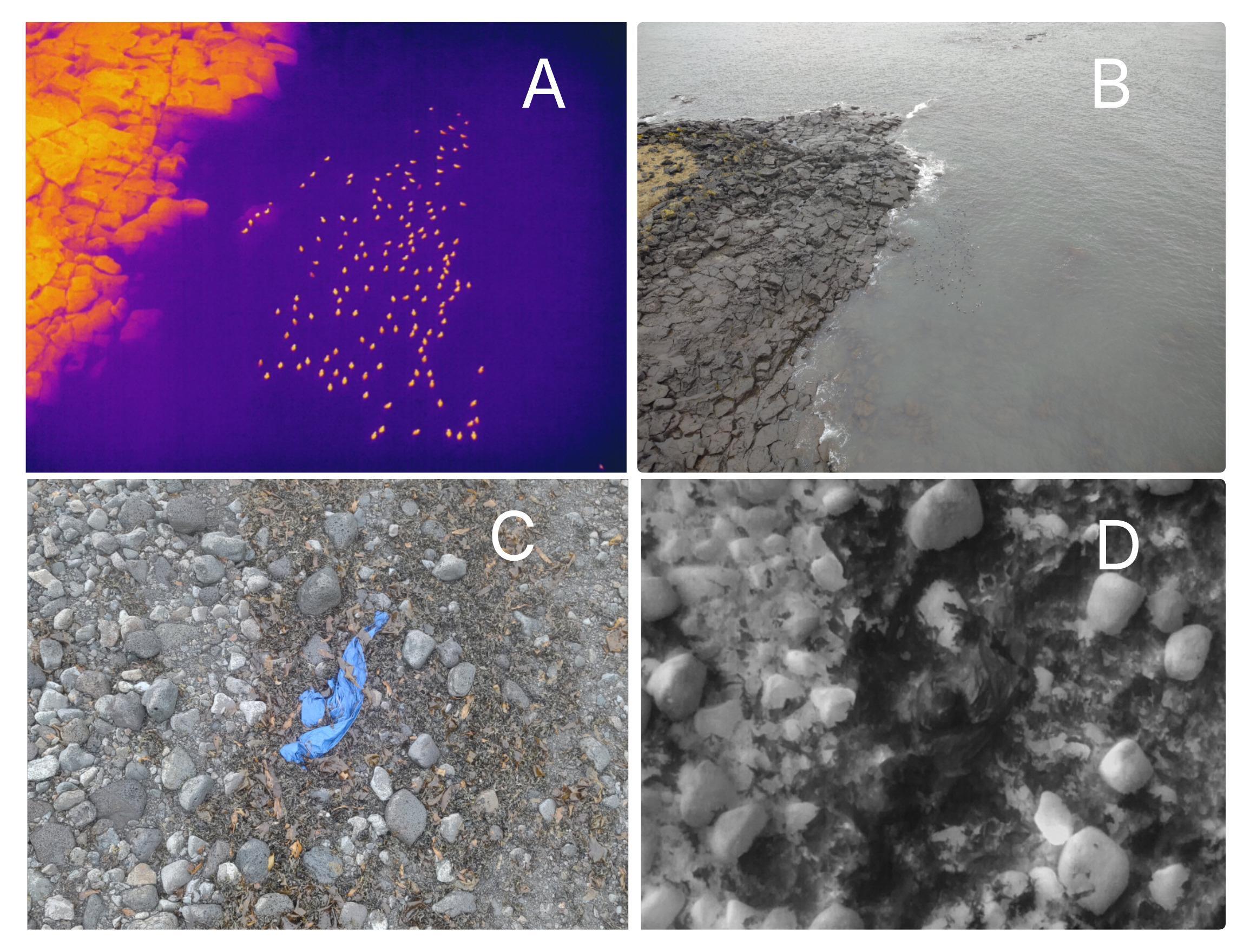}
\caption{RGB/IR image pairs from a 2025 QSAR operation near Mekoryuk, Alaska, illustrating limitations of thermal imagery. A-B (IR-RGB): Thermal FLIR sensors help spot features that may be hard to see in RGB imagery, but can also produce misleading detections, such as the
ducks shown in A.  C-D (RGB-IR): Some features, such as debris, are harder to detect in thermal wavelengths.}
\label{fig:rgb-ir}
\vspace{-5mm}
\end{figure*}

\section{Introduction}
\IEEEPARstart{W}{ilderness} Search and Rescue (WSAR) operations require teams of trained operators to locate indicators of missing persons, including man-made objects, tracks, and the persons themselves, across vast areas that can be difficult to access. Uncrewed aerial systems (UASs) offer a fast, maneuverable, and relatively inexpensive way to sense these environments, helping WSAR teams inspect terrain and identify locations that warrant closer investigation. Consequently, UASs have attracted growing interest among WSAR teams, with existing work exploring their integration into WSAR workflows and the use of visual and thermal sensing for indicator detection~\cite{drones4030038, schedl_2020,  BroylesHaynerEtAl2022, UAV_WSAR}. However, collecting aerial imagery is only part of the problem. WSAR teams must also determine where potentially relevant objects appear in the imagery, often under challenging environmental conditions and within time-sensitive operations. 

A typical WSAR workflow involves both aerial and terrestrial search. WSAR teams commonly use crewed and uncrewed aircraft to collect conventional visible (EO/RGB) and infrared (IR) imagery. RGB imagery provides intuitive situational awareness, while IR imagery can identify heat signatures indicative of human presence, or of humans themselves \cite{schedl_2020, BroylesHaynerEtAl2022}. In vast, remote areas where WSAR operations usually take place, UAS imagery helps ground teams execute search grids faster, extend awareness into hazardous terrain, and locate possible evacuation routes for critical patients \cite{drones4030038, BroylesHaynerEtAl2022}. These capabilities can reduce search time and improve responder safety.

\begin{figure}[th]
\centering\includegraphics[scale=0.9]{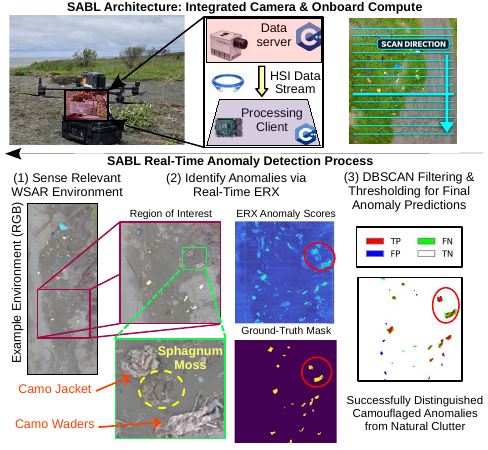}
\caption{SABL is a tightly integrated hardware/software edge computing pipeline that enables real-time, on-drone hyperspectral anomaly detection for WSAR.}   \label{fig:system_design}
\end{figure}

RGB and IR imagery, however,  have significant limitations. RGB imagery suffers from many of the same factors that challenge human vision: search environments are often cluttered, contain repetitive patterns that humans readily filter out, and may include brush or tree canopies that partially occludes objects that would otherwise be recognizable by shape \cite{liu2026rtrx}. Furthermore, missing persons may also wear camouflage, or earth-toned clothing that is difficult to distinguish from natural backgrounds in high-altitude aerial imagery (Figure \ref{fig:system_design}). IR imagery complements RGB sensing by measuring emitted thermal radiation, but cannot reliably identify materials that indicate human presence without a strong heat signature, such as tents, tarps, or discarded clothing. It may also fail to detect missing persons wearing insulated clothing, or deceased individuals~\cite{BroylesHaynerEtAl2022} (Figure \ref{fig:rgb-ir}).  Neither modality provides a direct way to differentiate anomalous materials that are partially occluded, deliberately camouflaged, insulated, or lacking a conventional thermal signature.

\begin{figure}[th]
\centering\includegraphics[scale=0.45]{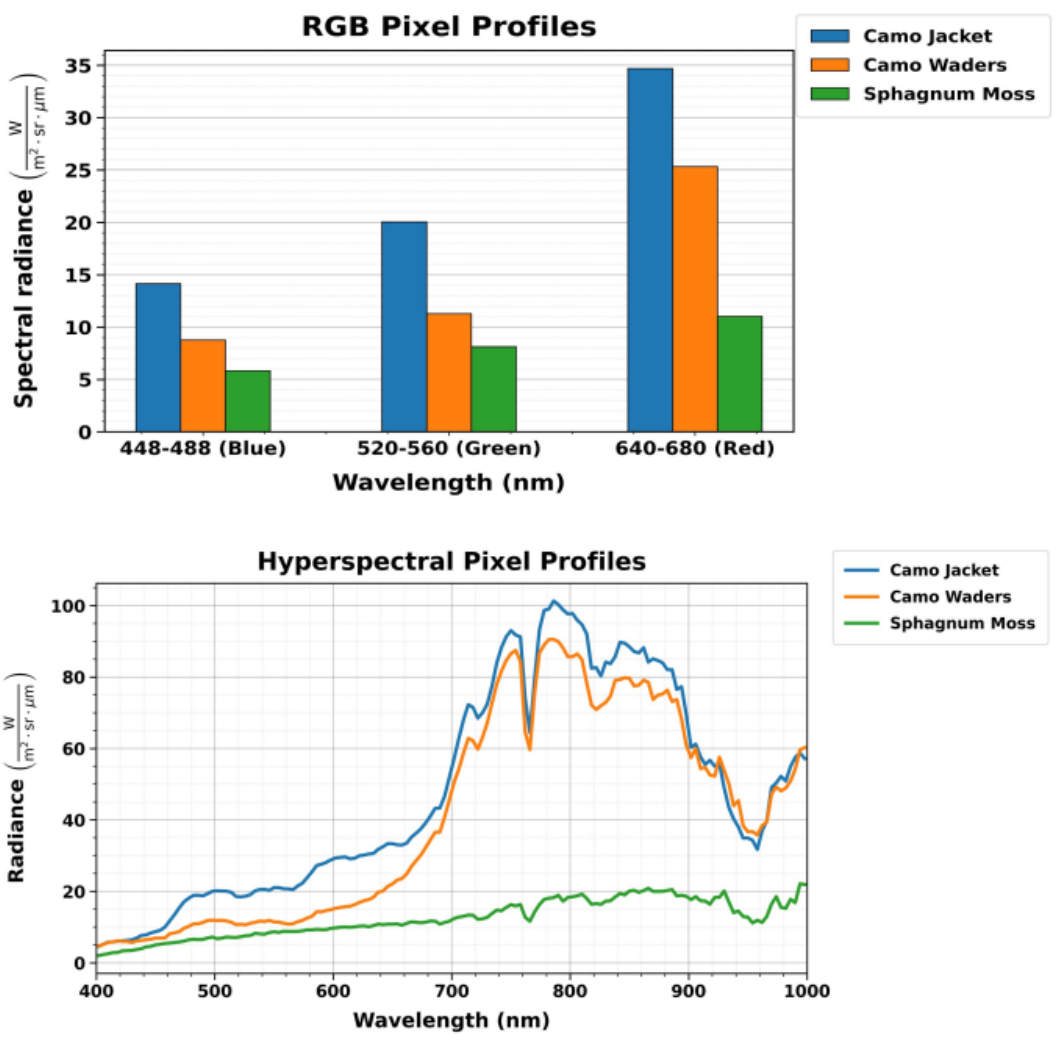}
\caption{Comparison of RGB versus HSI pixel profiles for camouflaged materials highlighted in Figure \ref{fig:system_design}. The RGB bands use the average radiance across the defined wavelength range. Notably, the HSI profile distinguishes the camouflaged wading gear from the neighboring moss, particularly at wavelengths beyond the red edge ($>700$ nm).}   \label{fig:profile}
\end{figure}

Hyperspectral imaging (HSI) offers a complementary source of information. Unlike RGB cameras, which capture three broad spectral channels, hyperspectral cameras measure spectral radiance across many (typically 100+) narrow spectral bands (Figure \ref{fig:profile}). These measurements can expose material-composition differences that are not apparent in conventional imagery, making HSI promising for detecting anomalous materials and objects in complex scenes~\cite{su2021hyperspectral, matteoli2010tutorial}. This additional information comes at a computational cost. Hyperspectral sensors generate large data volumes, and many UAS-compatible systems use push-broom sensors that acquire imagery one line at a time. Pixel-wise inference on these high-dimensional streams requires substantial computation, and subsequent geometric correction and photogrammetry introduce significant additional latency~\cite{gomez2008correction, ERX}. State-of-the-practice workflows store imagery for post-flight processing on high-performance computing equipment.

For time-sensitive WSAR operations, processing imagery as it is acquired can be as important as detecting anomalies accurately, because missing persons lost or injured may face hypothermia, hypovolemia, and shock if stranded. Onboard processing removes dependence on data offloading, thereby allowing potential indicators to be localized while a mission is underway. Achieving this capability, however, requires an integrated pipeline that can receive the sensor stream, decode hyperspectral measurements, and execute detection quickly enough to keep pace with acquisition on resource-constrained hardware.

Hyperspectral anomaly detection (HAD) provides a promising starting point. HAD algorithms identify pixels that differ statistically from a sensed background distribution~\cite{RX, ERX}. HAD is a common workload across agriculture, defense, environmental monitoring, etc., but is rarely adapted to real-time processing~\cite{goudemant2024onboard, shimoni2019hyperspectral, abdulsalam2024unsupervised}. HAD techniques are generally evaluated on corrected and post-processed data, and often run on entire datasets to identify global anomalies~\cite{RX}. Recent improvements in on-drone processing power and hyperspectral camera miniaturization have increased interest in real-time HAD. Prior works have demonstrated real-time hyperspectral detection of specific environmental phenomena~\cite{REMIX, spectramine}, near-real time on-drone HAD~\cite{watson2026real}, and an algorithm suited to real-time HAD for push-broom HSI sensors~\cite{ERX}. In our earlier work, we introduced Real-Time Reed-Xiaoli (RT-RX), an edge computing pipeline that integrates real-time HSI streaming with real-time HAD~\cite{liu2026rtrx}. RT-RX demonstrates that HSI can be acquired and processed in real-time onboard a UAS, but its detection outputs are limited to pixel-level anomaly scores. In a WSAR scenario, this places the burden of interpreting and filtering noisy HAD outputs on the operators, which may cause them to overlook or misinterpret critical information during time-sensitive searches.

In this work, we present Spectral Anomaly-Based Localization (SABL), an extension of RT-RX designed to generate interpretable outputs for WSAR operators (Figure \ref{fig:system_design}). SABL retains RT-RX's real-time hyperspectral streaming and line-based anomaly detection while adding spatial filtering via Density Based Spatial Clustering of Applications with Noise (DBSCAN)~\cite{DBSCAN}. DBSCAN suppresses isolated
detections and retains dense anomaly regions for operator inspection instead of raw pixel-wise detections.

To evaluate SABL, we worked with Quinhagak Search and Rescue (QSAR), a real WSAR team in remote Alaska, and conducted a 10-day data-collection campaign. We collected HSI of search-relevant objects in-situ, including clothing, fishing equipment, survival gear, and other objects which would immediately cue close inspection. We show that SABL processes hyperspectral data at $\sim$5.7 ms per line on a Raspberry Pi 4B, and achieves approximately 56\% and 99\% object-level precision on real WSAR search indicators across two realistic scenes. 

The main contributions of SABL are as follows:

\begin{itemize}

    \item \textbf{A real-time, on-drone hyperspectral edge-computing system for WSAR.} SABL's processing pipeline comprises HSI streaming, line-based HAD, and spatial filtering to identify and localize WSAR search indicators.
    
    \item \textbf{A field-collected hyperspectral dataset developed in collaboration with a real WSAR team in remote, rural Alaska.}  SABL provides a field-grounded  evaluation for HAD in austere environment, with a data collection guided by a real WSAR team.

    \item \textbf{A application-driven evaluation of system performance and detection capabilities.} We evaluate SABL's processing throughput relative to the sensor acquisition rate and characterize pixel- and object-level detection performance across two field scenes with differing background complexity. Our evaluation identifies both the conditions under which SABL produces useful search indicators and the environmental factors continue to challenge reliable detection.

\end{itemize}

\section{Background}
\label{sec:Background}

\subsection{Search and Rescue in Remote Alaska}

\begin{figure*}[!th]
    \centering
    \includegraphics[scale=0.9]{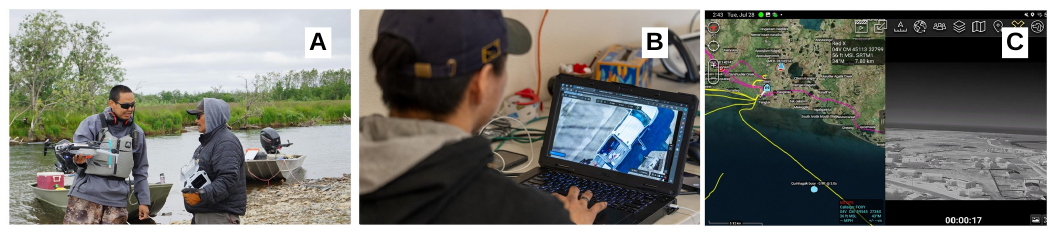}
    \caption{Images from recent SAR trainings in Quinhagak, Alaska. A) UAS operations are currently limited to small quadrotor aircraft equipped with EO/IR sensors. B) Yup'ik pilot Bryan Jones Jr. pilots a drone from Bethel, Alaska, during a Spring 2024 training. C) HSI inferences could be integrated with existing Quinhagak Search and Rescue (QSAR) infrastructure, such as the Android Team Awareness Kit (ATAK), to communicate actionable insights to field teams in real time.}
    \label{fig:qsar}
    \vspace{-5mm}
\end{figure*}

Quinhagak, Alaska, lies along the Bering Sea coast in the Yukon–Kuskokwim Delta, amid a landscape of rivers, lakes, and permafrost. It is one of 56 Yup'ik villages off the road system in western Alaska where villagers rely on snow machines, small marine skiffs, barges, and Cessna-style airplanes for transportation and logistics. Like many communities across the circumpolar north, daily life is typified by an emphasis on subsistence hunting, fishing, and gathering, which can result in a high operational tempo for locally led rescue missions. For instance, a village may experience up to three SAR missions per week during peak summer months, when villagers are regularly traversing the tundra to hunt and gather food. 

Villages in the Yukon-Kuskokwim delta, due to their remoteness, generally manage their own SAR processes. Quinhagak Search and Rescue (QSAR), is a non-profit organization which trains and organizes volunteer responders from the local community. When an event occurs, QSAR is generally activated by tribal public safety officers, who collect information about the missing person using a Lost Person Behavior survey \cite{LPB}. The survey documents the individual's intended travel plans, physical description (e.g., clothing, age, and vehicle type), and last known location to inform search strategies. Additionally, indigenous teachings, tracking skills, and understanding of their environment and human behavior form a traditional knowledge system that is central to QSAR.

QSAR operators also use small quadrotor UASs equipped with EO/IR sensors to quickly inspect hazardous or difficult-to-reach areas (e.g., coastal mudflats; Figure \ref{fig:qsar}). Missing persons in Quinhagak generally wear insulated or camouflaged clothing,  and carry earth-toned or camouflaged hunting and fishing equipment which often does not appear distinctly in EO/IR imagery (Figure \ref{fig:system_design}). We hypothesize that a  system capable of real-time HAD could enable this detection capability.

\subsection{Real-Time Hyperspectral Anomaly Detection: Algorithms and Systems}

 Hyperspectral anomaly detection (HAD) identifies unusual materials without requiring a predefined library of indicator signatures. The Reed-Xiaoli (RX) algorithm~\cite{RX} is a canonical approach that models the background of an HSI as a multivariate Gaussian distribution and calculates the Mahalanobis distance between each pixel's spectral vector and the estimated background. Pixels whose distance exceeds a user-set threshold are classified as anomalies. Because conventional RX estimates the background distribution across an entire image, it is not inherently suited to continuous, line-based processing. The exponentially moving Reed-Xiaoli (ERX) algorithm~\cite{ERX} addresses this limitation, using exponentially moving averages (EMAs) to adapt the background model to changing scenery while producing anomaly scores on a line-by-line basis in real time. To improve efficiency, ERX also employs sparse random projection (SRP) to reduce spectral dimensionality and Cholesky decomposition to avoid explicitly inverting the covariance matrix (Equation \ref{eq:maha}). 

Watson et al.~\cite{watson2026real} demonstrated near-real-time on-drone HAD using RX on batches of 1,000 lines. REMIX~\cite{REMIX} provides a streaming and decoding pipeline that connects a push-broom HSI camera and an onboard computer, enabling data to be transferred and processed as it is acquired. SpectraMine \cite{spectramine} exhibits real-time on-drone hyperspectral detection of camouflaged landmines using a pre-trained lightweight convolutional neural network (CNN) as well as spatial clustering. Real-Time Reed Xiaoli  (RT-RX)~\cite{liu2026rtrx}  integrates REMIX with ERX to achieve real-time on-drone HAD, but outputs noisy pixel-level detections, limiting its usability in WSAR. 

\section{Design}
As shown in Figure \ref{fig:design}, SABL extends the RT-RX edge computing pipeline~\cite{liu2026rtrx} by adding spatial localization. The underlying RT-RX architecture integrates push-broom HSI acquisition, REMIX-based streaming and decoding, and line-based ERX HAD to produce pixel-level anomaly scores onboard a UAS in real time. SABL retains this foundation, and introduces a DBSCAN-based spatial clustering stage that groups anomalous pixels into candidate search regions for operator inspection.   

\subsection{Real-Time Hyperspectral Data Acquisition}

\begin{figure*}[!t]
    \centering
    \includegraphics[scale=0.9]{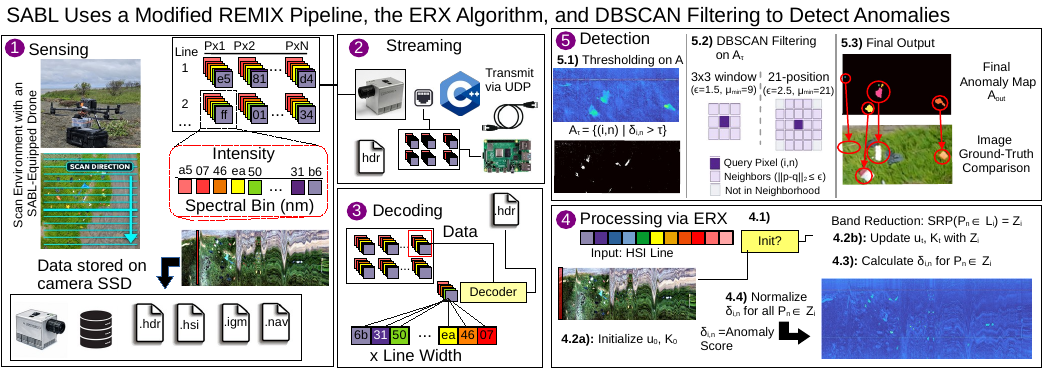}
    \caption{SABL uses a 5-stage pipeline to implement real-time hyperspectral anomaly detection for push-broom HSI sensors.}
    \label{fig:design}
    \vspace{-5mm}
\end{figure*}

SABL begins executing as the UAS senses its environment. Shown in Figure \ref{fig:design}(1), as the push-broom camera senses, it collects and stores lines of hyperspectral data in an internal storage device. Each line $L$ contains $N$ pixels, and each pixel $P$ contains $C$ channels, each of which stores measured upwelling radiance values for a fixed wavelength range (i.e., bin) of light as a short or long integer. These values are serialized by channel, pixel, and line, and are stored linearly in an ENVI-format file (\texttt{.hsi}) \cite{wolfe2018hyperspectral}. The encoding information of the file, such as data endianness, spectral bin wavelengths, bin width (i.e., spectral resolution), line width (i.e., the number of pixels in a line), line length (i.e., the total number of lines), and sampling rate are stored in a header file (\texttt{.hdr}). Other relevant information, including camera navigation data and automatically computed image geometry information, are stored alongside raw data to aid in photogrammetry.

As data is sensed, SABL transmits data from the camera to an onboard embedded device for processing. Data is transferred via a modified version of REMIX \cite{REMIX}. As shown in Figure \ref{fig:design}(2), SABL relies on a custom C++ data transmission server which executes within the hyperspectral camera. The server establishes a connection to a REMIX client on the embedded device, and transmits the header file to establish the data format. Both the client and server create line buffers, which are used to store intermediate HSI lines in transfer. As the header file is transmitted, the server begins continuously polling the HSI file from the camera's data drive. It reads raw bytes into the line buffer, and immediately transmits them to the embedded device via UDP. As the stream is received by the embedded device, pixels are entered into its line buffer. Once an entire line is available, SABL dispatches the line to the REMIX decoder and removes the line from the buffer. Once a data line is removed from the buffer, it is split into a 1-dimensional array of channel values organized in row-major order by pixel. This decoded line is then passed to ERX \cite{ERX} for detection. 

\subsection{Real-Time ERX Anomaly Detection}

The exponentially moving Reed-Xiaoli (ERX) algorithm \cite{ERX}, briefly described in Section \ref{sec:Background}, identifies anomalies by comparing the Mahalanobis distance $\delta$ of each pixel to a user-set  $\tau$. Equation \ref{eq:maha} shows Mahalanobis distance as calculated by ERX via Cholesky decomposition for a specific pixel $z_{i,n}$, where $i$ is the line index and $n$ is the pixel index in that line. $z_{i,n}$ is compared to the most up-to-date mean $u_t$ and covariance $K_t$. $T_t$ is the lower triangle of $K_t^{-1} + 10^{-5}I$.  

\begin{equation}
    \label{eq:maha}
    \delta_{i,n} = \sqrt{[(z_{i,n}-\mu_t)T_t^{-1}][(z_{i,n}-\mu_t)T_t^{-1}]^T}
\end{equation}

SABL implements ERX as a function pointer within REMIX. SABL uses a modified REMIX client written in C++ running on the UAS's onboard computer. As each line is decoded, it is sent to ERX, which is implemented in Python, via \texttt{pybind11}. As ERX processes each line, it returns the normalized Mahalanobis distance for each pixel, which is the raw anomaly score.

\subsection{Density-Based Spatial Filtering}
\label{sec:DBSCAN}

A pixel is classified as anomalous by ERX when $\delta_{i,n} > \tau$. The resulting set of anomalous pixels is denoted by $A_{\tau}$. Each detected pixel is represented by its location in the hyperspectral image: $A_{\tau} = \{(i,n) \:|\: \delta_{i,n} > \tau\}$. SABL also associates detected pixels with their corresponding UAS-sensed locations using navigation and geometric information provided by the camera. 

There are several reasons why pixel-level anomaly scores do not necessarily correspond to physical anomalous objects. An individual pixel may contain spectral information from multiple materials, known as spectral mixing~\cite{hong2026hyperspectral}, which is particularly relevant near object boundaries. ERX can also produce weak or fragmented responses within a physical object. Moreover, isolated anomalies may arise from sensor artifacts, inconsistent illumination, or local background heterogeneity. Presenting every anomalous pixel independently can therefore produce a large number of spatially scattered detections, many of which may be false positives, which can undermine WSAR missions by diverting scarce personnel and resources to false alarms, potentially delaying assistance to people in need.  

Based on the intuition that dense clusters of anomalous pixels are more likely to be interesting to WSAR operators than scattered spectral noises, SABL uses Density-Based Spatial Clustering of Applications with Noise (DBSCAN)~\cite{DBSCAN} to detect anomalous \textit{locations} as collections of densely clustered anomalous pixels. SABL only retains pixels in these \textit{locations}, filtering all others as noises. Given the anomalous pixel set $A_\tau$, the $\epsilon$-neighborhood of a pixel $p$ is defined as 

\begin{equation}
    \mathcal{N}{\epsilon}(p) = \{q \in A_{\tau} : ||q-p||_2 \leq \epsilon\}. 
\end{equation}

A pixel is considered a \textit{core pixel} when its neighborhood contains at least $\mu_{\min}$ anomalous pixels:

\begin{equation}
    |\mathcal{N}{\epsilon}(p)| \geq \mu_{\min}. 
\end{equation}

DBSCAN groups spatially proximate anomalous pixels into clusters based on their local density. Pixels that cannot be assigned to a cluster are treated as noise and excluded from the filtered output. SABL uses Euclidean distance between anomalous pixel coordinates in the HSI. 

The spatial density requirements determine the degree of filtering. 
We use two DBSCAN configurations as exemplars: low-radius and high-radius. Low-radius filtering $(\epsilon=1.5, \mu_{min} = 9)$, captures a central pixel and its entire 8-connected neighborhood. This assures that core pixels are surrounded by detected anomalies, limiting exposure to spectral mixing and illumination inconsistencies. High-radius filtering $(\epsilon=2.5, \mu_{min} = 21)$, extends the neighborhood to include the central pixel and 20 surrounding locations within a $5\times5$ window. Because DBSCAN uses Euclidean distance, the four corners of the window fall outside the radius and thus are not considered. Therefore, all 21 locations within this neighborhood must contain detected anomalies for a pixel to qualify as a core pixel. The two configurations impose complete-neighborhood density requirements over different spatial extents. The smaller neighborhood favors compact neighborhoods, whereas the larger neighborhood requires density over a larger region, potentially suppressing smaller or more fragmented responses. 

Our density filtering requirement is motivated by WSAR needs. Positive detections help operators prioritize locations for closer investigation. False alarms caused by isolated spectral outliers or debris can consume vital search time and resources. Filtering, however, can also suppress true positives when spectral mixing or limited indicator visibility produce sparse responses. This introduces a tradeoff between recall and precision. 

As SABL runs, a bounded intermediate buffer $A_{buf}$ groups incoming thresholded ERX predictions into batches and retains overlapping raw prediction lines to preserve spatial neighborhood information across batch boundaries. A halo of $2\lfloor\epsilon\rfloor$ rows on each side provides sufficient neighboring context. Once sufficient future context is available, SABL finalizes batches of 100 rows and writes them sequentially to the final output buffer $A_{out}$, retaining only the raw context needed for subsequent batches. At acquisition completion, the remaining rows are finalized by treating locations beyond the image boundary as background.

SABL's ultimate outputs are shown in Figure \ref{fig:design}(5). They include an anomaly map $A_{out}$, where predicted background pixels are black, and a colored map of raw $\delta$ values for each pixel, representing $A$, which can be used for anomalous object detection. SABL also reports the GPS position of the center-most point in the cluster as determined by the cameras navigation information.

\section{Implementation}

\subsection{Field Deployment in Quinhagak, Alaska}
\begin{figure}[!t]
\centering
\includegraphics[scale=1]{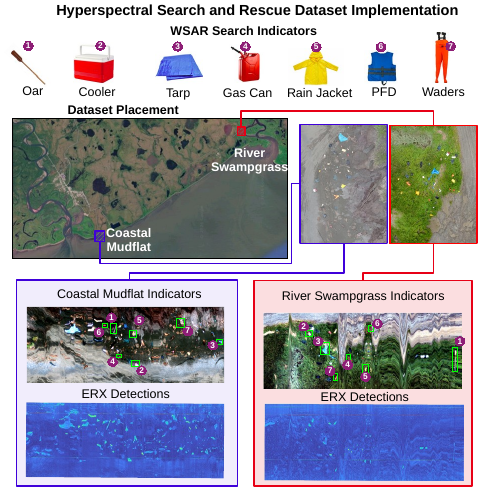}
\caption{We collected two hyperspectral scenes of dozens of relevant QSAR indicators in the subarctic tundra surrounding Quinhagak, Alaska.}
\label{fig:imp}
\end{figure}

During water rescues, QSAR personnel typically identify items that often float after accidents or wash up on beaches or riverbanks. To simulate a QSAR mission scenario, we arranged relevant items (mostly donated by the community), including clothing (waders, jackets, boots), fishing and marine equipment (nets, life jackets, buckets, buoys), and outdoor recreation gear (tarps, backpacks, storage canisters), in two settings: a mudflat on the coast of the Bering sea, and a patch of swamp grass near the Arolik river. Figure ~\ref{fig:imp}, shows each location around Quinhagak, and overhead hyperspectral scans of both scenes constrained to RGB using the 440nm, 520nm, and 640nm channels as blue, green, and red respectively. The Mudflat scene contains significant environmental clutter, including rocks, driftwood, and sphagnum moss. The Swamp Grass scene has a more uniform background material composition, but provides challenging textural and reflective diversity. 

\subsection{System Setup and Configuration}

We performed imaging using a DJI M350 RTK equipped with a Corning microHSI 410 SHARK hyperspectral camera, which captures 150 bands in VNIR (400-1000 nm) with a spectral bin width of 4 nm. The camera automatically performs radiometric correction and dark subtraction on the HSI before it gets transmitted. Each HSI line consists of 682 pixels, each with 150 channels, and each channel stores measured upwelling radiance values as a 32-bit integer. The camera's sensing rate is configured to 100 Hz. Our modified REMIX streaming pipeline consists of 436 lines of C++ code. We heavily modified ERX python code from prior work~\cite{ERX} to integrate it with REMIX via \texttt{pybind11}. We executed this system on a Raspberry Pi 4B connected to the camera via a Cat6 Ethernet cable. The microHSI 410 SHARK, Raspberry Pi, networking hardware, an RGB camera (GoPro Hero 11), and a 12V lithium-ion battery were packaged in waterproof casing and affixed to the drone before sensing.

\subsection{Dataset}
As shown in Figure ~\ref{fig:imp}, our dataset contains raw, unprocessed hyperspectral scanlines. To preserve SABL's real-time processing workflow, we apply no additional corrections. The dataset contains 75,419 scanlines: 44,953 lines from the the Mudflat scene and 30,466 from the Swamp Grass scene. Accompanying drone telemetry includes position, attitude, altitude, ground speed, and timing information. Flights were conducted on July 17--18, 2026, under overcast conditions, at an approximate speed of 2 km/h and an altitude of 100 ft above ground level (AGL), yielding a ground sampling distance (GSD) of 2.13 cm. Each scene includes multiple overhead passes. We selected one pass from each scene for manual annotation in ENVI, using RGB footage as a visual reference. The resulting annotated subset comprises 4700 scanlines (Mudflat: 2,900, Swamp Grass: 1,800), and is used to evaluate SABL in Section \ref{sec:Results}. This dataset is openly available: https://zenodo.org/records/24382506. 

\subsection{Hyperparameter Settings}
Both ERX and DBSCAN contain tunable hyperparameters that would affect SABL's output. For our evaluation, SABL follows ERX's original published configurations~\cite{ERX}, using 5 SRP dimensions and a momentum rate of 0.1 to update Gaussian background statistics. Buffer size is set to 0, meaning detection starts immediately as soon as sensing starts. SABL does, however, tune the threshold for each scene as shown in Figure \ref{fig:maps_1} and Tables \ref{tab:Pixel} and \ref{tab:Object}. For DBSCAN, we use the configurations described in Section \ref{sec:DBSCAN} for low- and high- radius filtering.

\section{Evaluation and Results}
\label{sec:Results}

\subsection{Runtime Performance}

\begin{figure}[!t]
    \centering
    \includegraphics[scale=0.87]{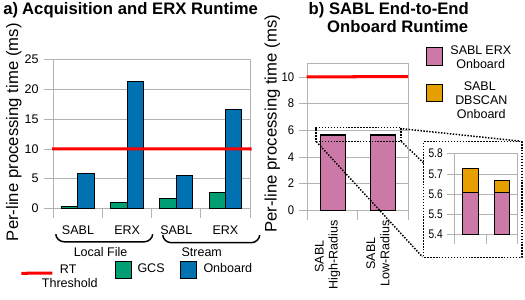}
    \caption{SABL runtime results. a) SABL can process HSI data in real-time on a small embedded device. b) SABL's DBSCAN post-processing increases runtime by 1-2\%.} 
    \label{fig:results}
    \vspace{-5mm}
\end{figure}

Figure~\ref{fig:results}(a) compares SABL and ERX in terms of average per-line processing time across four configurations combining two execution contexts and two data-access methods. For a direct comparison with standalone ERX, we measure the runtime of SABL's two components: ERX and DBSCAN filtering, separately
(Figure~\ref{fig:results}(b)). The two execution contexts are 1) an embedded system onboard a UAV, and 2) a ground control station (GCS), which is a moderately powerful laptop used for high-level flight planning. For each of these contexts, we consider execution on both a local file, and an HSI data stream. Each of these scenarios has a use case in WSAR:

\begin{itemize}
    \item \textbf{(Onboard, Streaming):} Streaming directly to an onboard embedded device is the principal real-time use case and the goal of SABL. 

    \item \textbf{(GCS, Streaming):} Streaming to GCS hardware is useful in rendering data offline. In scenarios where not all data can be used online by the UAV, real-time GCS re-rendering can allow WSAR workers to quickly access anomaly detection data and make search decisions in the field.

    \item \textbf{(Onboard, Local File):} UAVs running online anomaly detection may need to save data that overruns buffers for offline processing during downtime or dispatch.

    \item  \textbf{(GCS, Local File):} This case is a baseline and state of the practice, where data is offloaded from the UAV and processed on the GCS.
\end{itemize} 

\begin{figure*}[ht]
\centering
\includegraphics[scale=0.55]{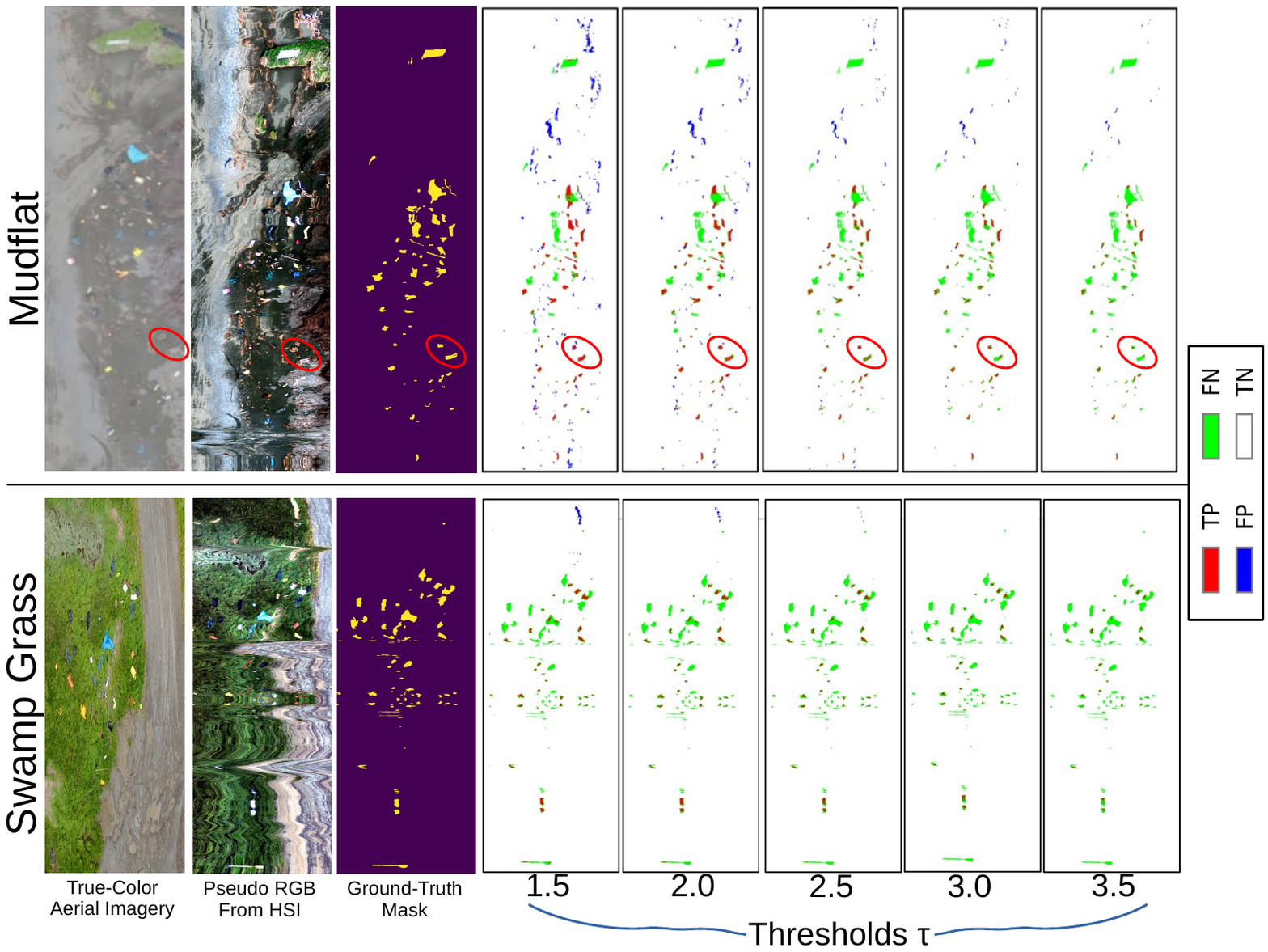}
\caption{Qualitative visualization of the detection maps for both scenes across a range of thresholds $\tau$. Corresponding quantitative metrics are presented in Tables \ref{tab:Pixel} and \ref{tab:Object}.}
\label{fig:maps_1}
\end{figure*}

Real-time performance necessitates a line processing time of 10 ms or less, denoted by the red lines in Figure \ref{fig:results}, to meet the camera's 100-Hz acquisition rate. SABL maintains real-time across all 4 execution contexts, even achieving real-time processing on a resource-scarce Raspberry Pi 4B when both streaming and reading local HSI files, achieving line processing times of 5.6 ms and 5.9 ms respectively. The SOTA ERX implementation can also achieve real-time processing, but only on the GCS platform. On the embedded device, ERX achieves average line processing times of 21.34 ms and 16.61 ms for local files and streams respectively, violating the real-time constraint. On the Raspberry Pi, SABL outperforms ERX by $\sim$3x for stream processing and $\sim$3.6x for local file processing. On the GCS, SABL outperforms ERX by a factor of $\sim$1.5x for stream processing, and $\sim$3.7x for local file processing. Notably, both SABL and ERX experience slower processing times for local files as opposed to streams on the embedded device. This is due to the device using an SD card for storage, whose bandwidth is lower than the stream established by our on-camera software.

Figure ~\ref{fig:results}(b) shows SABL's performance when DBSCAN filtering is included. As lines are processed via ERX, SABL places them into an intermediate $A_{buf}$ which is processed at 1 Hz and appended to the output buffer. This process takes an average of 125.02 $\mu s$ for high-radius DBSCAN and an average of 65.03 $\mu s$ for low-radius DBSCAN per line. These results show that spatial filtering is an extremely light-weight addition to ERX anomaly detection. 

\begin{table*}[ht]
\caption{Pixel-level detection results for both scenes from $\tau$=1.5 to $\tau$=3.5. We report the mean and standard deviation over 5 runs with random seeds: 0-4.}
\label{tab:Pixel}
\centering
\begin{tabular}{c|*{3}{c}|*{3}{c}}
\multirow{2}{*}{\bf Threshold}  &\multicolumn{3}{c|}{\bf Mudflat} &\multicolumn{3}{c}{\bf Swamp Grass} \\ &Precision(\%) &Recall(\%) &F1-Score(\%) &Precision(\%) &Recall(\%) &F1-Score(\%)  \\ \hline 
1.5 &$42.67 \pm 2.69$  &$41.23 \pm 0.78$ &$41.89 \pm 1.35$ &$86.67 \pm 9.53$  &$21.75 \pm 1.82$ &$34.56 \pm 1.56$ \\
2.0        
&$53.69 \pm 3.52$  &$33.26 \pm 0.59$ &$41.04 \pm 1.17$ &$94.61 \pm 5.75$  &$17.03 \pm 1.84$ &$28.76 \pm 2.38$ \\
2.5
&$63.09 \pm 3.04$  &$26.69 \pm 0.36$ &$37.49 \pm 0.38$ &$98.19 \pm 2.17$  &$13.45 \pm 1.71$ &$23.61 \pm 2.60$ \\
3.0 
&$69.23 \pm 2.41$  &$19.84 \pm 0.65$ &$30.82 \pm 0.64$ &$99.78 \pm 0.28$  &$9.96 \pm 1.83$ &$18.07 \pm 3.03$ \\
3.5  &$76.25 \pm 3.18$  &$14.86 \pm 0.58$ &$24.86 \pm 0.68$ &$99.92 \pm 0.16$ &$7.03 \pm 1.61$ &$13.10 \pm 2.81$ 
\end{tabular}
\end{table*}

\begin{table*}[ht]
\caption{Object-level detection results for both scenes from $\tau$=1.5 to $\tau$=3.5. We report the mean and standard deviation over 5 runs with random seeds: 0-4.}
\label{tab:Object}
\centering
\begin{tabular}{c|*{3}{c}|*{3}{c}}
\multirow{2}{*}{\bf Threshold}  &\multicolumn{3}{c|}{\bf Mudflat} &\multicolumn{3}{c}{\bf Swamp Grass} \\ &Precision(\%) &Recall(\%) &F1-Score(\%) &Precision(\%) &Recall(\%) &F1-Score(\%)  \\ \hline 
1.5     &$17.97 \pm 0.63$ &$69.84 \pm 1.67$ &$28.57 \pm 0.81$ &$65.16 \pm 16.87$ &$38.62 \pm 2.78$ &$47.41 \pm 2.89$ \\
2.0        
&$23.44 \pm 2.48$ &$65.25 \pm 0.66$ &$34.42 \pm 2.61$ &$85.72 \pm 10.62$ &$35.40 \pm 3.81$ &$49.66 \pm 3.08$ \\
2.5
&$32.22 \pm 1.85$ &$61.31 \pm 1.97$ &$42.17 \pm 1.17$ &$92.15 \pm 8.34$ &$32.64 \pm 3.89$ &$47.81 \pm 3.22$ \\
3.0 
&$43.07 \pm 2.65$ &$58.36 \pm 1.67$ &$49.49 \pm 1.67$ &$\mathbf{98.64 \pm 1.67}$ &$29.20 \pm 4.34$ &$44.85 \pm 5.12$ \\
3.5 &$\mathbf{56.01 \pm 4.55}$ &$55.08 \pm 0.80$ &$55.43 \pm 1.90$ &$95.33 \pm 6.09$ &$25.29 \pm 5.34$ &$39.75 \pm 6.96$
\end{tabular}
\end{table*}

\subsection{Anomaly Detection Performance}

Tables \ref{tab:Pixel} and \ref{tab:Object} show the pixel- and object-level quantitative performance, respectively, of SABL on the two scenes across a range of thresholds $\tau$. Table \ref{tab:Pixel} shows a clear relationship across both datasets between $\tau$ and the pixel-level metrics: as also demonstrated in Figure \ref{fig:maps_1}, increasing $\tau$ generally reduces false positive pixels (blue) but increases false negatives (green); i.e., leads to higher pixel-level precision but lower recall. In the Mudflat scene, pixel-level precision increases from 42.67\% to 76.25\, while recall decreases from 41.23\% to 14.86\%. This effect is more noticeable in the Swamp Grass scene, where precision increases from 86.67\% to 99.92\%, and recall decreases from 21.75\% to 7.03\%. Pixel-level $F_1$-score decreases as $\tau$ increases in both scenes. This demonstrates that higher $\tau$'s produce more selective anomaly masks, but recover smaller fractions of the ground-truth WSAR indicators at the pixel level.

Pixel-level metrics do not fully capture the role of SABL in WSAR. SABL is intended to group anomalous pixels via DBSCAN as objects which are then relayed to operators. To quantify object-level detection, we extract predicted regions by connected-component labeling of the DBSCAN-filtered anomaly mask. Additionally, to accommodate fragmented detections, which may stem from variability in illumination and surface roughness (e.g., folds in tarps or clothing), we allow multiple predicted regions to contribute to the detection of a single ground-truth object. Each predicted region is associated with the ground-truth object for which it achieves the greatest 
ground-truth-normalized overlap, defined as
\begin{equation}
    c_{p,g} = \frac{|P_p \cap G_g|}{|G_g|},
\end{equation}
where $P_p$ and $G_g$ denote the pixel sets of predicted region $p$ and ground-truth object $g$, respectively. An object is considered detected if the combined coverage reaches at least 5\% of its pixels. This permissive criterion prioritizes identifying search cues rather than perfect delineation of an anomalous object. Predicted regions with no ground-truth overlap, or associated with objects whose combined coverage
falls below 5\%, are counted individually as false positives. Ground-truth objects that do not meet the coverage criterion are counted as false negatives. This metric tolerates fragmentation, where a single ground-truth object may be represented by a set of small, disjoint, but highly anomalous regions. 

Object-level results show different threshold-specific behavior than pixel-level results. As displayed in Table \ref{tab:Object}, in the more cluttered Mudflat scene, object-level $F_1$-score increases from 28.57\% to 55.43\%; precision improves from 17.97\% to 56.01\%, while recall decreases more modestly (from 69.84\% to 55.08\%) than at the pixel level. The more uniform Swamp Grass scene exhibits slightly different behavior. Object-level precision reaches 98.64\% at $\tau=3$, but object-level $F_1$-score is higher at $\tau=2$, with a precision of 85.72\%, and recall of 35.4\%. Increasing the threshold beyond $\tau=2$ increases precision at the cost of recall, showing that the threshold that creates the most selective detections does not strike the strongest balance between precision and recall for less cluttered scenes.  

Together, these results show that the relationship between $\tau$ and detection quality depends on both the background environment and system goal. Pixel-level $F_1$-score favors lower $\tau$, but object-level detections that can be more useful for WSAR operators benefit from relatively higher $\tau$. Additionally, qualitative inspection also shows that higher $\tau$ can suppress detections of camouflaged objects
(see the red circles in Figures~\ref{fig:system_design} and~\ref{fig:maps_1}); therefore a higher $\tau$ is not always ideal as it may weaken a key motivation for hyperspectral sensing: revealing targets that are difficult to
discern in RGB imagery.  

\subsection{Ablation Study: DBSCAN Filtering}

\begin{figure*}[ht]
\centering
\includegraphics[scale=0.55]{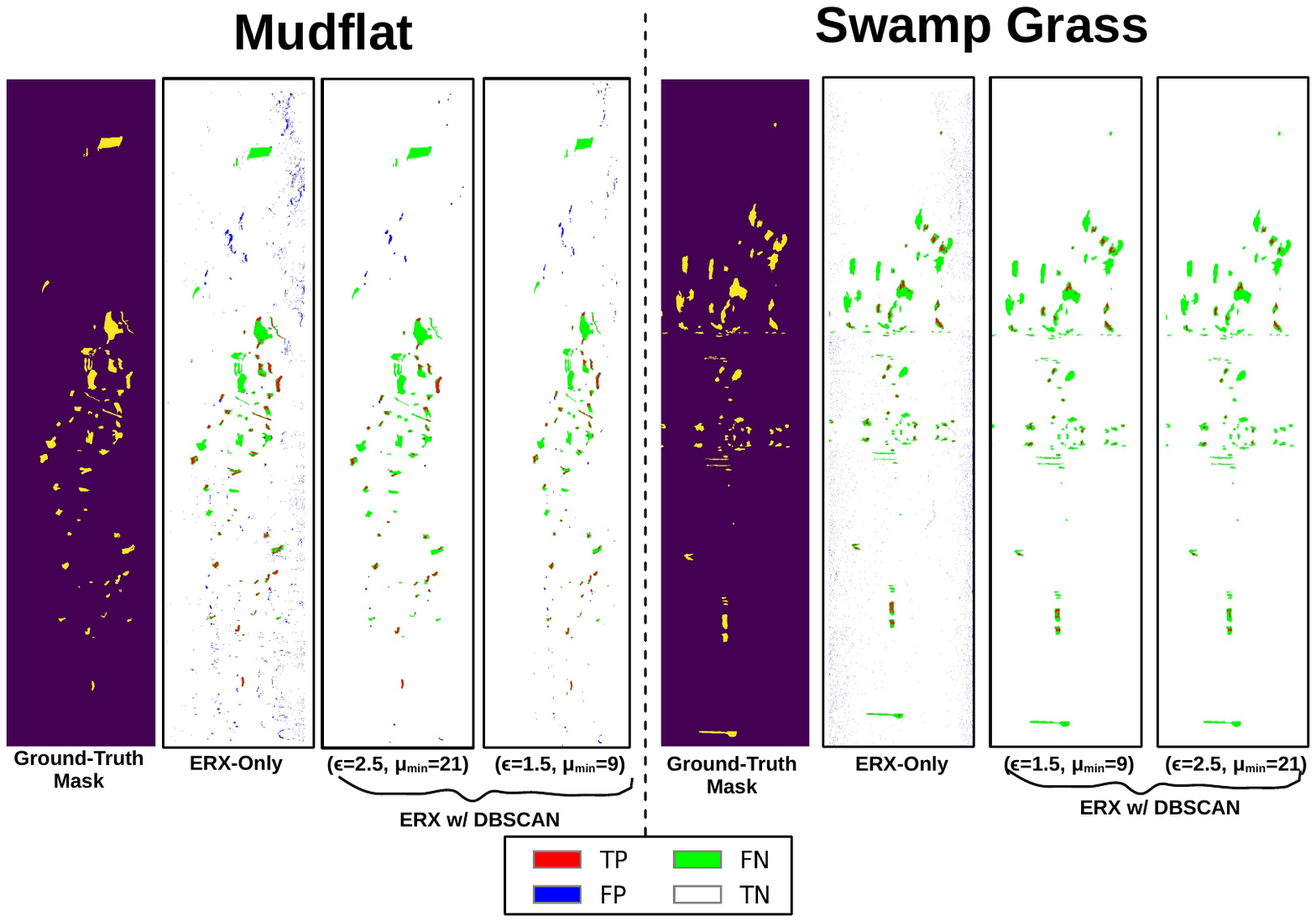}
    \caption{Qualitative visualization of the detection maps across both scenes for the ERX anomaly detector, with and without low- and high-radius DBSCAN filtering. Corresponding quantitative metrics are presented in Tables \ref{tab:ERX-ablation} and \ref{tab:DBSCAN-ablation}.}
    \label{fig:dbscan}
\end{figure*}

\begin{table*}[ht]
\caption{Pixel- and object-level results across both scenes for the ERX anomaly detector, with and without DBSCAN filtering. We report the mean and standard deviation over 5 runs with random seeds: 0-4.}
\label{tab:ERX-ablation}
\centering
\begin{tabular}{c|*{3}{c}|*{3}{c}}
\multirow{2}{*}{\bf Scene}  &\multicolumn{3}{c|}{\bf ERX} &\multicolumn{3}{c}{\bf ERX w/ DBSCAN Filtering} \\ &Precision(\%) &Recall(\%) &F1-Score(\%) &Precision(\%) &Recall(\%) &F1-Score(\%)  \\ \hline 
Mudflat / Pixel 
&$37.98 \pm 1.52$  &$22.98 \pm 0.90$ &$28.61 \pm 0.77$ &$76.25 \pm 3.18$  &$14.86 \pm 0.58$ &$24.86 \pm 0.68$ \\
Mudflat / Object 
&$14.96 \pm 1.85$  &$71.80 \pm 1.23$ &$24.71 \pm 2.56$ &$56.01 \pm 4.55$ &$55.08 \pm 0.80$ &$55.43 \pm 1.90$ \\
\hline
Swamp Grass / Pixel        
&$37.66 \pm 1.88$ &$14.40 \pm 1.99$ &$20.71 \pm 2.01$ &$99.78 \pm 0.28$  &$9.96 \pm 1.83$ &$18.07 \pm 3.03$ \\
Swamp Grass / Object        
&$75.82 \pm 4.96$ &$43.68 \pm 5.99$ &$55.09 \pm 4.83$ &$98.64 \pm 1.67$ &$29.20 \pm 4.34$ &$44.85 \pm 5.12$ 
\end{tabular}
\end{table*}

\begin{table*}[ht]
\caption{Pixel- and object-level results across both scenes for low- and high-radius DBSCAN filtering. We report the mean and standard deviation over 5 runs with random seeds: 0-4.}
\label{tab:DBSCAN-ablation}
\centering
\begin{tabular}{c|*{3}{c}|*{3}{c}}
\multirow{2}{*}{\bf Scene}  &\multicolumn{3}{c|}{\bf $\mathbf{(eps=1.5, \mu_{min} = 9)}$} &\multicolumn{3}{c}{\bf $\mathbf{(eps=2.5, \mu_{min} = 21)}$} \\ &Precision(\%) &Recall(\%) &F1-Score(\%) &Precision(\%) &Recall(\%) &F1-Score(\%)  \\ \hline 
Mudflat / Pixel    
&$61.71 \pm 2.92$ &$18.45 \pm 0.87$ &$28.37 \pm 0.83$ &$76.25 \pm 3.18$  &$14.86 \pm 0.58$ &$24.86 \pm 0.68$  \\
Mudflat / Object    
&$29.39 \pm 2.51$  &$68.33 \pm 1.49$ &$41.01 \pm 2.20$ &$56.01 \pm 4.55$ &$55.08 \pm 0.80$ &$55.43 \pm 1.90$ \\
\hline
Swamp Grass / Pixel        
&$99.78 \pm 0.28$  &$9.96 \pm 1.83$ &$18.07 \pm 3.03$ &$100.00 \pm 0.00$ &$7.45 \pm 1.55$ &$13.83 \pm 2.68$ \\
Swamp Grass / Object        
&$98.64 \pm 1.67$ &$29.20 \pm 4.34$ &$44.85 \pm 5.12$ &$98.82 \pm 2.35$ &$22.07 \pm 3.20$ &$35.99 \pm 4.31$ 
\end{tabular}
\end{table*}

As shown in Figure \ref{fig:dbscan} and Table \ref{tab:ERX-ablation}, we qualitatively and quantitatively assess the contribution of DBSCAN filtering to the ERX anomaly detector. For each scene, we use the threshold which achieves the highest object-level precision, as highlighted in bold in Table \ref{tab:Object} (Mudflat: $\tau$=3.5, Swamp Grass: $\tau$=3). Table \ref{tab:ERX-ablation} demonstrates that DBSCAN filtering substantially increases detection precision across both scenes. In the Mudflat scene, pixel-level precision increases from 37.98\% to 76.25\%, while object-level precision increases from 14.96\% to 56.01\%. We also see an increase in object-level $F_1$-score from 24.71\% to 55.43\%. This indicates that DBSCAN filtering improves the balance between false positive and true positive detections in the more cluttered Mudflat environment, despite filtering out some genuine search indicators. 

The more uniform Swamp Grass scene exhibits a different response. DBSCAN filtering increases object-level precision from 75.82\% to 98.64\%; however, object-level $F_1$-score declines from 55.09\% to 44.85\%. This can be possibly explained by the fact that ERX already produces relatively accurate object-level detections, and DBSCAN further filters out both true and false positives. We conclude that while DBSCAN improves precision significantly, its effect on object-level detection quality depends on background characteristics.

Table ~\ref{tab:DBSCAN-ablation} shows the influence of filtering neighborhood size on detection performance. In the more cluttered Mudflat scene, the larger neighborhood attains better object-level precision and $F_1$-score. With a higher-radius filter, small, background-adjacent non-anomalies like driftwood pieces that can appear anomalous to ERX can be filtered out. By contrast, in the more consistent Swamp Grass scene, the smaller neighborhood provides stronger performance. The larger neighborhood provides a marginal precision improvement (0.18\%), as precision is already significantly high. It therefore filters out mainly true positives, thus reducing $F_1$-score through lower recall. These results show that neighborhood-size selection influences the precision-recall tradeoff, and that a single DBSCAN configuration may not be suitable for all environments. 

\section{Limitations and Future Work}

While SABL demonstrates the feasibility of real-time, line-based hyperspectral anomaly detection for WSAR, several limitations must be addressed before operational use. First, object-level recall remains limited: at the reported operating points, SABL detects 55.08\% of annotated objects in Mudflat and 29.20\% in Swamp Grass. Spatial filtering can suppress true detections alongside false-positives, highlighting the need to balance false-alarm reduction against missed search indicators. SABL should, for this reason, be viewed as a tool for augmenting existing WSAR methods through identifying candidate search indicator, instead of a replacement for established WSAR workflows.

Second, our evaluation is limited to two scenes in the subarctic tundra, leaving SABL's performance in other WSAR environments
uncharacterized. Future work will evaluate the pipeline across a
broader range of terrains, weather conditions, vegetation cover,
illumination, and search-indicator materials to assess its
generalizability.

Finally, additional engineering efforts may improve onboard processing efficiency. We selected the Raspberry Pi 4B to evaluate SABL under tight computational constraints. Future work will investigate
additional embedded platforms, such as the NVIDIA Jetson devices, and characterize their processing latency. We also plan to extend the current partial C++
implementation of ERX to the full ERX processing path and measure the resulting performance gains.

\section{Conclusion}
We present SABL, an edge computing pipeline for on-drone \underline{S}pectral \underline{A}nomaly-\underline{B}ased \underline{L}ocalization, which performs line-by-line hyperspectral streaming, anomaly detection, and spatial filtering in real time. We demonstrate that SABL processes hyperspectral data at 5.67-5.73 ms per line on a Raspberry Pi 4B, exceeding the throughput required for whole-stream processing on a 100-Hz push-broom camera. SABL is evaluated on two field-collected scenes developed in collaboration with Quinhagak Search and Rescue (QSAR), a WSAR team in remote Alaska. Our results exhibit that spatial filtering can substantially improve object-level precision, particularly in cluttered environments. Our findings establish the feasibility of real-time on-drone hyperspectral processing for WSAR, while highlighting the importance of scene-dependent filtering. Future work will focus on adaptive filtering, further optimizing runtime performance, and integrating SABL into real human-in-the-loop WSAR workflows.


%

\section*{Acknowledgment}
This work was funded in part by the SUNY Research Foundation and the AFRL Mid-Atlantic Regional Hub (Award FA8750-22-2-0500) and was
performed with equipment donated by Corning Inc. We also thank the Quinhagak Search and Rescue team, Nalaquq LLC, and  Qanirtuuq Inc. for their help and support with this research. 

\bibliographystyle{IEEEtran}
\bibliography{ref.bib}

\end{document}